\documentclass[11pt]{article}

\usepackage[preprint]{acl}

\usepackage{times}
\usepackage{latexsym}
\usepackage{amsmath}
\usepackage{amsfonts}
\usepackage{graphicx,subcaption,multirow,wrapfig,pdfpages, booktabs}

\usepackage[T1]{fontenc}

\usepackage[utf8]{inputenc}

\usepackage{microtype}

\usepackage{inconsolata}

\usepackage{graphicx}

\title{Plasticity vs. Rigidity: The Impact of Low-Rank Adapters on Reasoning on a Micro-Budget}

\author{
  Zohaib Khan \and
  Omer Tafveez \and
  Zoha Hayat Bhatti \\
  University of Michigan \\
  \texttt{zohaibkh@umich.edu, omertaf@umich.edu, zohakh@umich.edu}
}

\begin{document}
\maketitle
\begin{abstract}
Recent advances in mathematical reasoning typically rely on massive scale, yet the question remains: can strong reasoning capabilities be induced in small language models ($\leq1.5\text{B}$) under extreme constraints? We investigate this by training models on a single A40 GPU (48GB) for under 24 hours using Reinforcement Learning with Verifiable Rewards (RLVR) and Low-Rank Adaptation (LoRA). We find that the success of this ``micro-budget" regime depends critically on the interplay between adapter capacity and model initialization. While low-rank adapters ($r=8$) consistently fail to capture the complex optimization dynamics of reasoning, high-rank adapters ($r=256$) unlock significant plasticity in standard instruction-tuned models. Our best result achieved an impressive 40.0\% Pass@1 on AIME 24 (an 11.1\% absolute improvement over baseline) and pushed Pass@16 to 70.0\%, demonstrating robust exploration capabilities. However, this plasticity is not universal: while instruction-tuned models utilized the budget to elongate their chain-of-thought and maximize reward, heavily math-aligned models suffered performance collapse, suggesting that noisy, low-budget RL updates can act as destructive interference for models already residing near a task-specific optimum.
\end{abstract}

\section{Introduction}

Reasoning tasks---such as mathematical problem solving, logical inference, and symbolic manipulation---remain among the most challenging domains for language models (LLMs). 
While scaling model size has historically improved reasoning ability \cite{wei2023chainofthoughtpromptingelicitsreasoning,openai2024gpt4technicalreport}, recent work suggests that sheer parameter count is not the only path forward. 
Methods such as reinforcement learning with verifiable rewards (RLVR) \cite{shao2024deepseekmathpushinglimitsmathematical,deepscaler2025} and supervised fine-tuning on structured reasoning traces \cite{muennighoff2025s1simpletesttimescaling,ye2025limoreasoning} have demonstrated that models can acquire advanced reasoning capabilities when guided by structured feedback and verifiable signals. 
However, the majority of these advances rely on large-scale models trained with extensive compute budgets, leaving open the question: 
how efficiently can small or mid-sized models be trained to reason well under tight computational constraints?

Recent studies point toward several promising directions for ``reasoning on a budget''. 
First, compact instruction-tuned models have shown latent reasoning potential that can be unlocked with a small number of high-quality examples---the so-called LIMO hypothesis \cite{ye2025limoreasoning} that fine-tuning quality matters more than quantity. 
Second, \citeauthor{muennighoff2025s1simpletesttimescaling} demonstrated that with as few as 1,000 curated problems and careful test-time control (methods such as ``budget forcing''), a 32B model can match or exceed proprietary systems in mathematical reasoning. 
Third, DeepScaleR extended reinforcement learning to long-context reasoning, showing that a 1.5B model can surpass much larger baselines by progressively increasing reasoning length during RL training \cite{deepscaler2025}. 
Together, these findings highlight a growing recognition that data curation, reward structure, and inference compute may be more decisive than raw scale.

Despite this progress, the literature still lacks a systematic study of how {parameter-efficient finetuning (PEFT)} methods interact with RLVR in small-model settings.
Most RL works employ full-parameter updates, assuming abundant GPU memory and stable optimization dynamics.
In contrast, parameter-efficient strategies such as Low-Rank Adaptation (LoRA) \cite{hu2021loralowrankadaptationlarge} offer a practical means to explore the trade-off between trainable capacity and reasoning performance.
Recent work has demonstrated that LoRA can be surprisingly expressive even under tight budgets: \citeauthor{schulman2025lora} showed that, up to a certain data-to-parameter ratio, LoRA finetuning can match or exceed full-parameter finetuning, provided the adapter placement and rank are tuned appropriately.
Similarly, the Tina series of models \cite{wang2025tinatinyreasoningmodels} achieved strong reasoning performance---reaching over 43\% Pass@1 on AIME24---by applying reinforcement learning with LoRA adapters to a 1.5B base model, at a fraction of the cost of full-scale training.
These results suggest that low-rank updates are not merely a compute-saving heuristic but can, under the right conditions, unlock reasoning behavior comparable to much larger or fully finetuned models.

However, how these dynamics extend to scenarios with \emph{extreme} computational constraints remains an open question.
Our work investigates the limits of reasoning optimization under a strict ``micro-budget'': \textbf{a single A40 GPU (48GB) restricted to 24 hours of training} (equating to approximately 7.2 USD\footnote{As per \url{vast.ai}---just a bit more than a cup of coffee}).
Under such tight constraints, where models may undergo fewer than 300 update steps, the interaction between the base model's initialization and the LoRA adapter's capacity becomes critical.
We explore this across a diverse set of small language models ($\leq1.5\text{B}$), including general instruction-tuned models, including those specialized for math and intensive reasoning.
By varying LoRA ranks ($r \in \{8, 64, 256\}$) within an RLVR framework using Group Relative Policy Optimization (GRPO), we test whether high-rank adapters can induce plasticity in small models even with minimal compute.

Our results reveal a stark dichotomy in how models respond to cheap post-training.
We find that generalist instruction-tuned models (and even the RL-tuned DeepScaleR) exhibit high \emph{plasticity}: when equipped with high-rank adapters ($r=256$), they rapidly learn to elongate their reasoning chains and maximize reward, significantly boosting performance on benchmarks like MATH500 and AIME24.
In contrast, heavily specialized models like Qwen2.5-Math-1.5B and Qwen3-0.6B display \emph{rigidity}: the noisy, low-budget RL updates act as destructive interference, causing performance collapse rather than refinement.
Ultimately, we propose that the most efficient path to reasoning on a budget is not to refine experts, but to catalyze generalists with high-rank adaptation.

\section{Methodology}

To investigate the limits of reasoning optimization under strict compute constraints, we adopted a parameter-efficient reinforcement learning framework. All experiments were conducted on a single NVIDIA A40 GPU (48GB VRAM) with a strict 24-hour training cutoff.

\subsection{Models}
We selected a diverse set of small language models ($\leq1.5\text{B}$ parameters) to evaluate how different initialization strategies affect plasticity under low-budget RL. Our selection spans three categories: models like {(1) Qwen2.5-1.5B-Instruct and (2) Llama-3.2-1B-Instruct} possessing broad knowledge but lacking specific reasoning optimization; models like {(3) Qwen2.5-Math-1.5B and (4) Qwen3-0.6B} with extensive pre-training or alignment for mathematics; and an RL-optimized benchmark like {(5) DeepScaleR-1.5B-Preview} to test whether "cheap" RL can further refine an already optimized policy.

\subsection{Datasets}

We utilized the {Open-RS} dataset \cite{dang2025reinforcementlearningreasoningsmall}, a collection of ~7000 reasoning problems containing diverse mathematical and logical queries.

For evaluation, we tracked model validation performance during training with {MATH500} and then did a final evaluation with the best rank/checkpoint on {AIME24/25 and AMC23} which are competition-level math problems.

\subsection{Training Procedure}
We implemented our training pipeline using the \texttt{verl} framework \cite{Sheng_2025}.

\paragraph{Fine-tuning with LoRA.}

\begin{figure*}[h]
    \centering
    \includegraphics[width=0.4\linewidth]{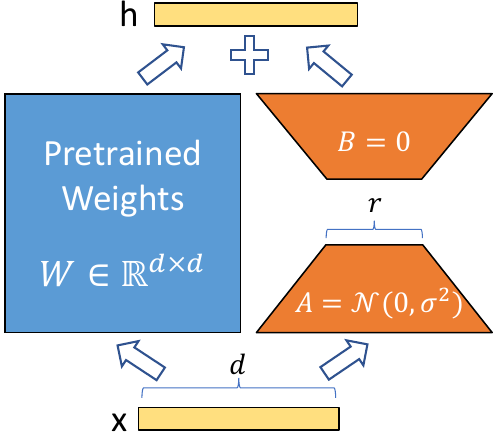}
    \caption{Low-Rank Adaptation (LoRA) mechanism. By optimizing only the low-rank matrices $A$ and $B$, we significantly reduce memory usage while retaining the ability to learn task-specific features.}
    \label{fig:lora}
\end{figure*}

Given the 48GB memory constraint, full-parameter fine-tuning was infeasible. We utilized {Low-Rank Adaptation (LoRA)} \cite{hu2021loralowrankadaptationlarge}, which freezes the pre-trained weights $W$ and injects trainable rank decomposition matrices $A$ and $B$, such that $W' = W + BA$, where $A \in \mathbb{R}^{r \times d}, B \in \mathbb{R}^{d \times r}$ (see Figure~\ref{fig:lora}).
We swept the rank $r \in \{8, 64, 256\}$ to test the hypothesis that higher ranks are necessary to capture the complex gradient updates of RLVR.

\paragraph{RLVR with GRPO.}
We employed {Group Relative Policy Optimization (GRPO)} \cite{shao2024deepseekmathpushinglimitsmathematical}, a policy gradient method designed for efficiency. Unlike PPO, which requires a memory-intensive value network, GRPO estimates the baseline from a group of $k$ sampled outputs for the same prompt.
\begin{itemize}
    \item \textbf{Rollouts:} We used a group size of $k=8$ to fit within the A40's memory.
    \item \textbf{Token Limit:} Following the configuration in \citeauthor{wang2025tinatinyreasoningmodels}, we capped the maximum response length at {3584 tokens} to encourage the generation of detailed chain-of-thought reasoning without exceeding context windows. Note that this is \textit{much} smaller than what other works like \citeauthor{deepscaler2025} use.
\end{itemize}

\paragraph{Reward Structure.}
We utilized a deterministic, verifiable reward function $R_{\text{total}}$:
\[
    R_{\text{total}} = 0.2 \cdot R_{\text{format}} + R_{\text{accuracy}}
\]
where $R_{\text{format}}$ provides a small shaping signal for adhering to the \texttt{<think>...</think>} structure, and $R_{\text{accuracy}}$ is a binary reward ($+1$) awarded solely if the final boxed answer matches the ground truth.

\section{Experimental Results}

We analyze the training dynamics and final performance of the five models to characterize the behavior of RLVR under strict compute constraints.

\subsection{Evolution of Training Reward}
\begin{figure*}[t]
    \centering
    \begin{subfigure}{0.32\textwidth}
        \includegraphics[width=\linewidth]{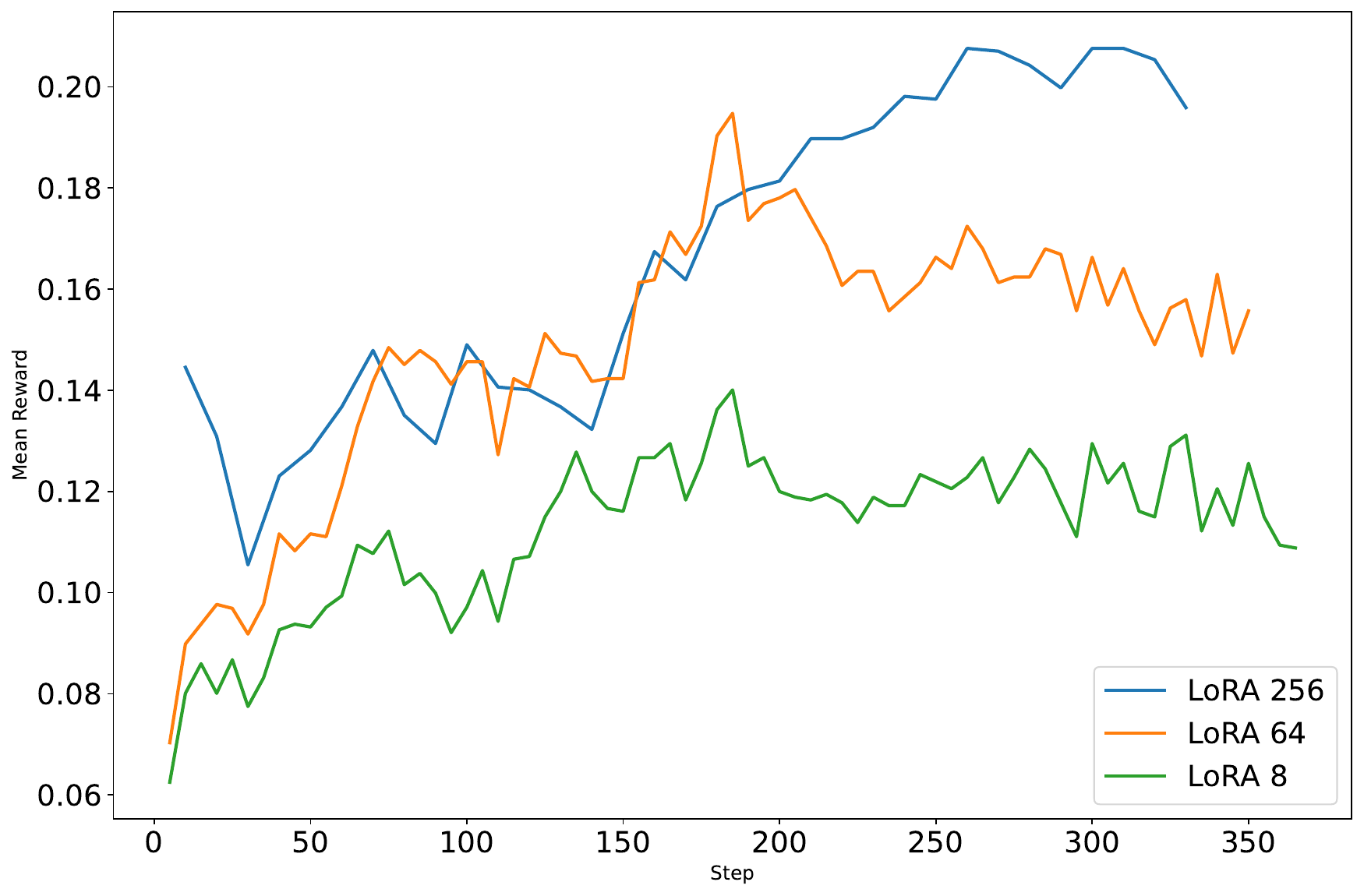}
        \caption{Qwen2.5-1.5B-Instruct}
    \end{subfigure}
    \hfill
    \begin{subfigure}{0.32\textwidth}
        \includegraphics[width=\linewidth]{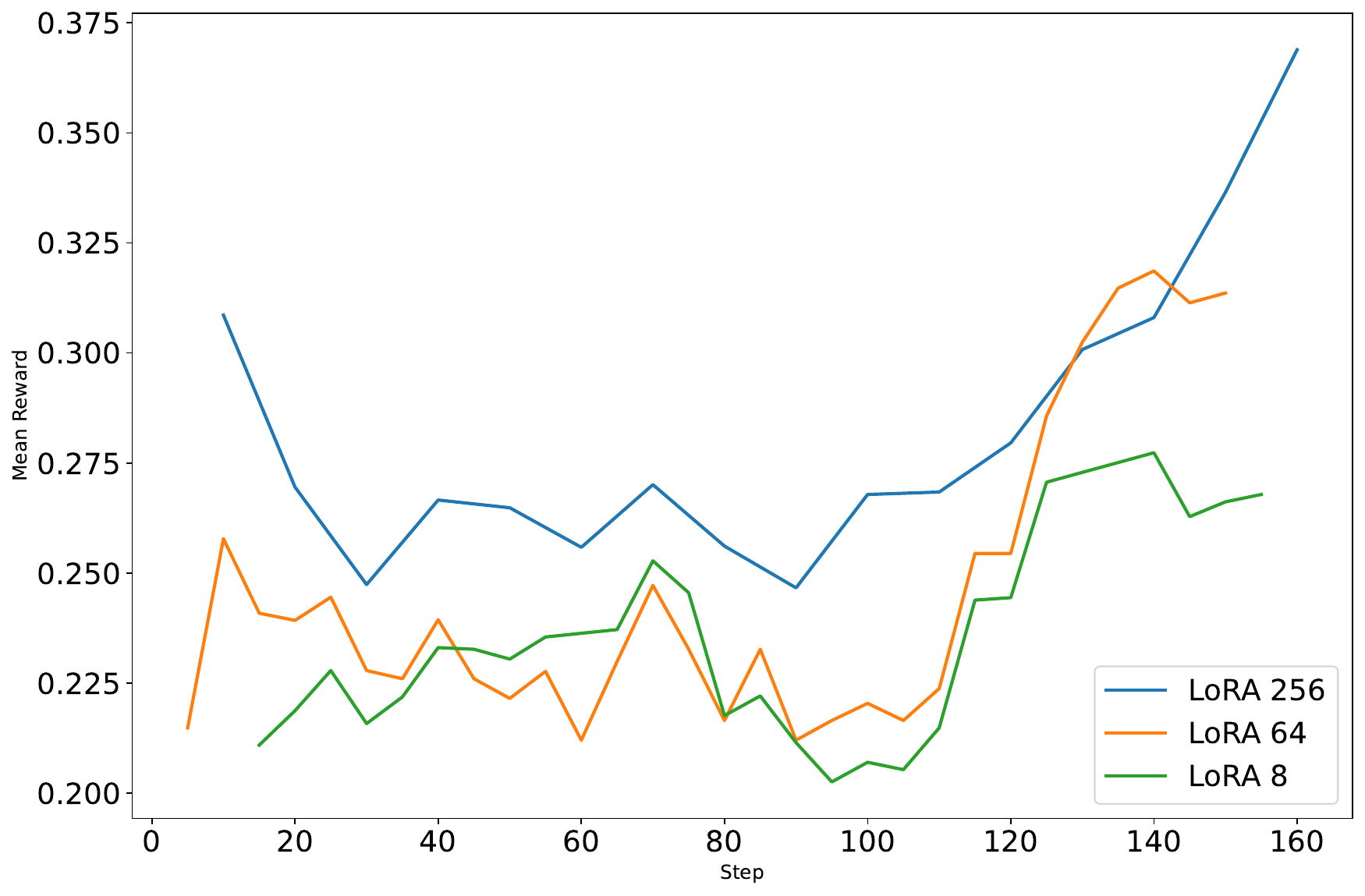}
        \caption{DeepScaleR-1.5B}
    \end{subfigure}
    \hfill
    \begin{subfigure}{0.32\textwidth}
        \includegraphics[width=\linewidth]{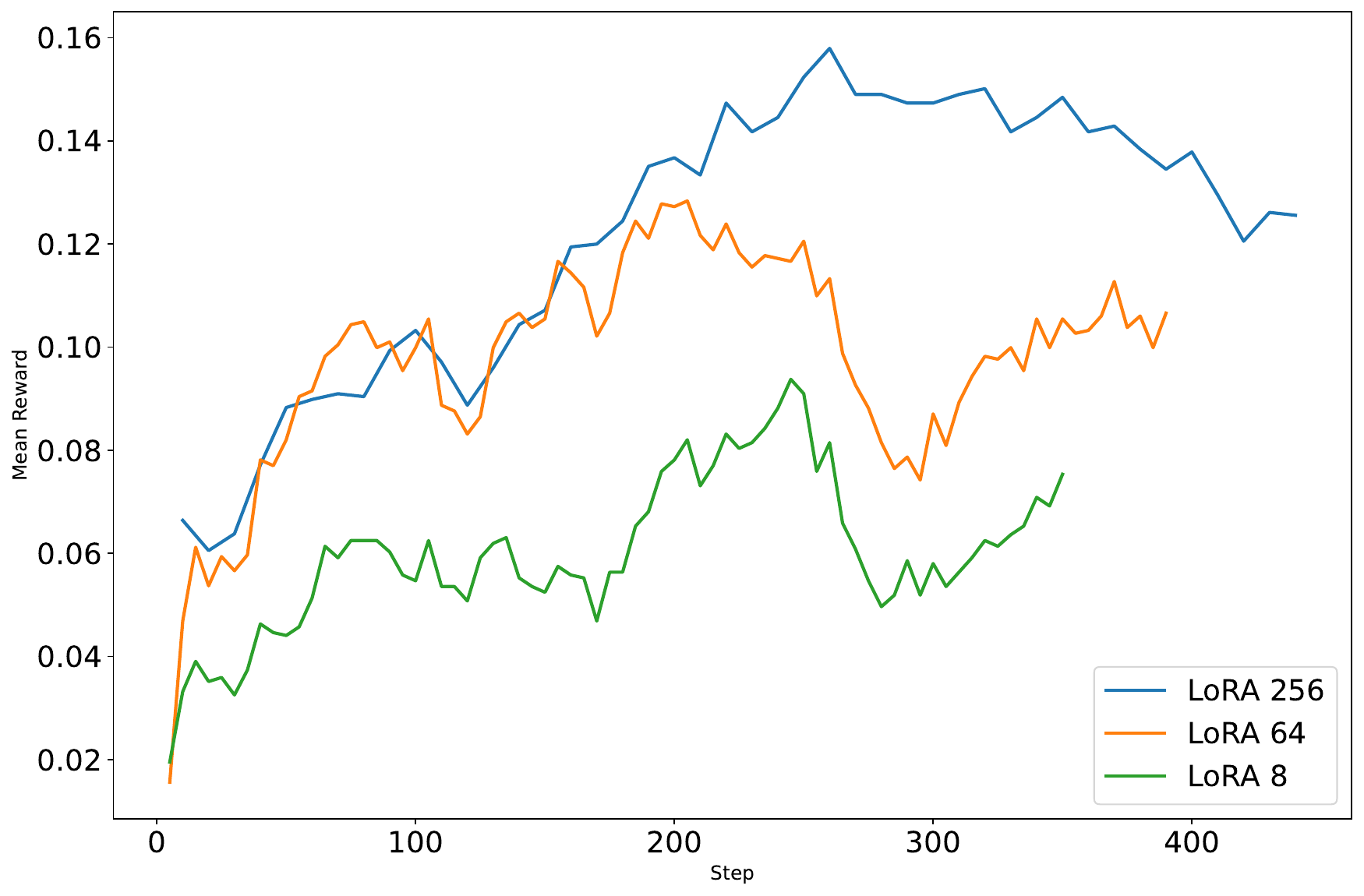}
        \caption{Llama-3.2-1B}
    \end{subfigure}
    
    \vspace{0.5em}
    
    \begin{subfigure}{0.32\textwidth}
        \centering
        \includegraphics[width=\linewidth]{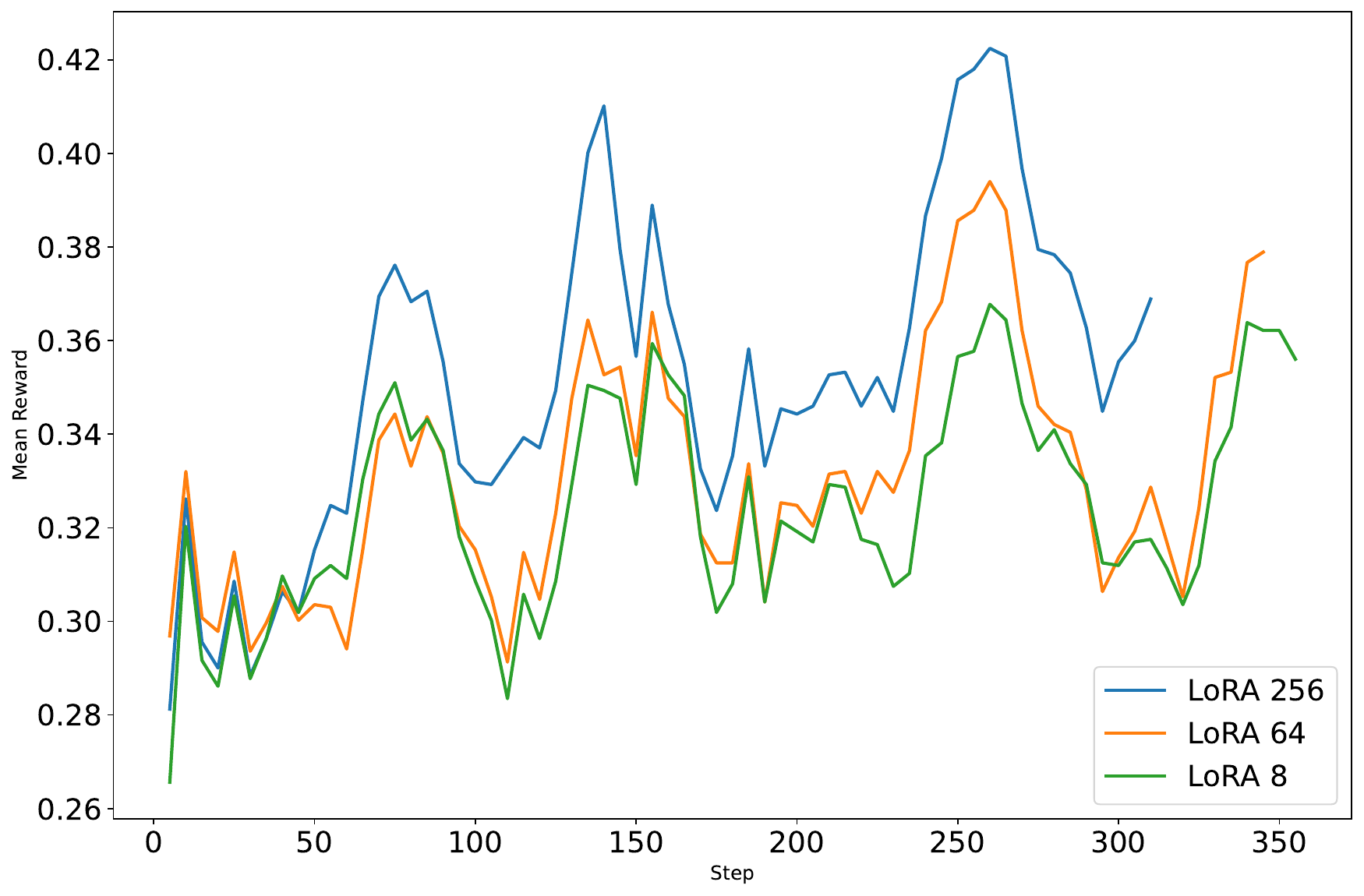}
        \caption{Qwen2.5-Math-1.5B}
    \end{subfigure}
    \hspace{1em}
    \begin{subfigure}{0.32\textwidth}
        \centering
        \includegraphics[width=\linewidth]{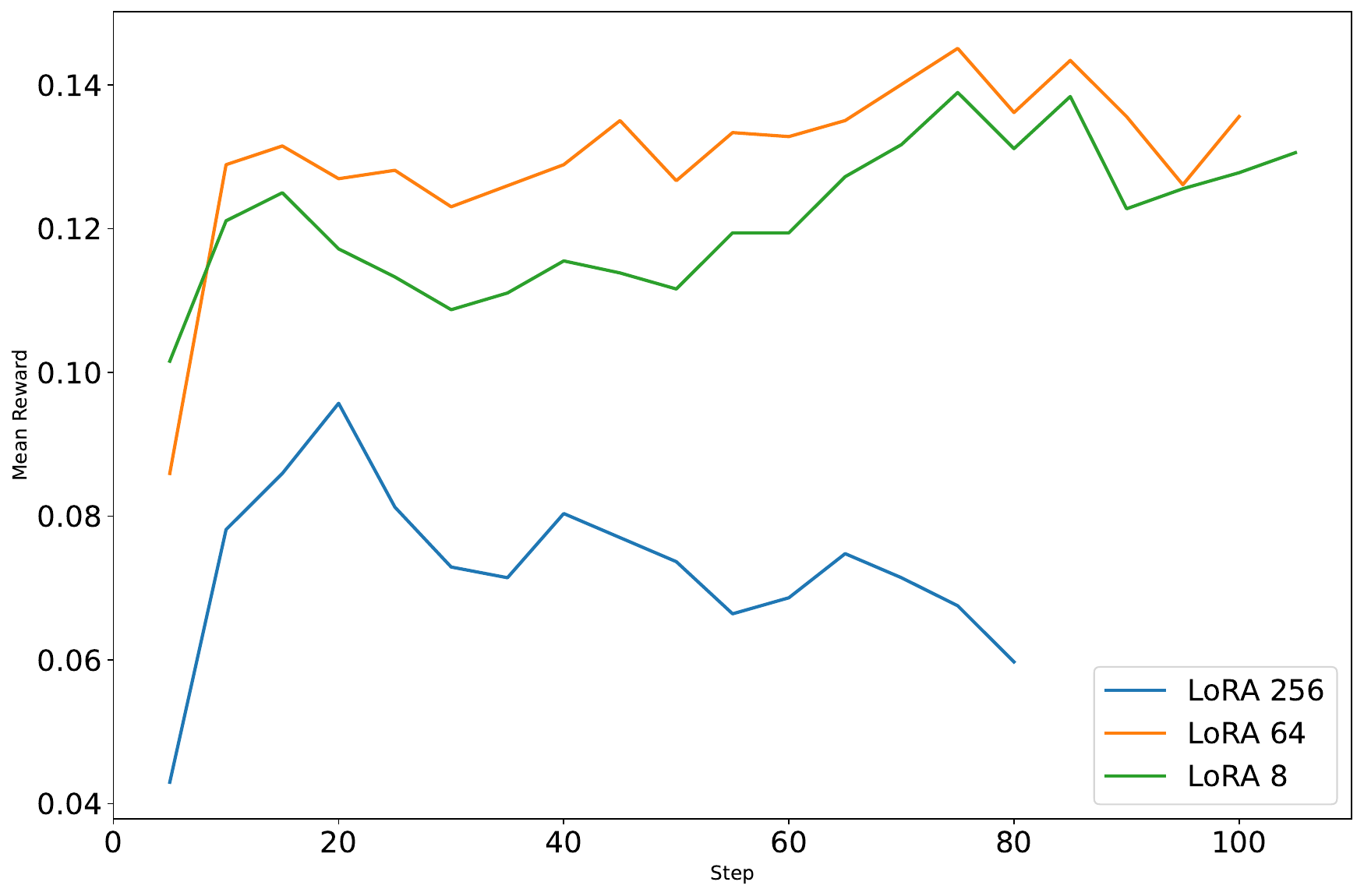}
        \caption{Qwen3-0.6B}
    \end{subfigure}
    
    \caption{\textbf{Evolution of Mean Reward.} High-rank adapters ($r=256$, blue lines) drive consistent learning for generalist models (Top Row), whereas models that underwent less conventional training (Bottom Row) struggle to optimize the reward signal.}
    \label{fig:rewards}
\end{figure*}

We first examine the ability of different models to optimize the verifiable reward signal (correctness + format) within the 24-hour budget.
As shown in Figure \ref{fig:rewards}, a clear distinction emerges based on adapter rank:
\begin{itemize}
    \item \textbf{Generalist Plasticity:} Qwen2.5-1.5B-Instruct and DeepScaleR-1.5B exhibit (almost) monotonic reward growth at high ranks ($r=256$) compared to lower ones. The high-rank adapters provide sufficient capacity to internalize the RL signal, aligning with what \citeauthor{schulman2025lora} found. Even for a model like Llama that isn't suited for reasoning, it too benefits hugely from this cheap training scheme, going from near-zero to double digits in the train reward.
    \item \textbf{Specialist Instability:} Qwen2.5-Math-1.5B shows significant instability at high ranks. Rather than converging, the reward signal fluctuates and degrades, suggesting the updates are conflicting with the model's pre-optimized manifold. An even more concerning result is how Qwen3-0.6B borderline collapses at higher rank updates, an exaggerated case of the previous model.
\end{itemize}

\subsection{Validation Performance (MATH500)}
\begin{figure*}[t]
    \centering
    \begin{subfigure}{0.32\textwidth}
        \includegraphics[width=\linewidth]{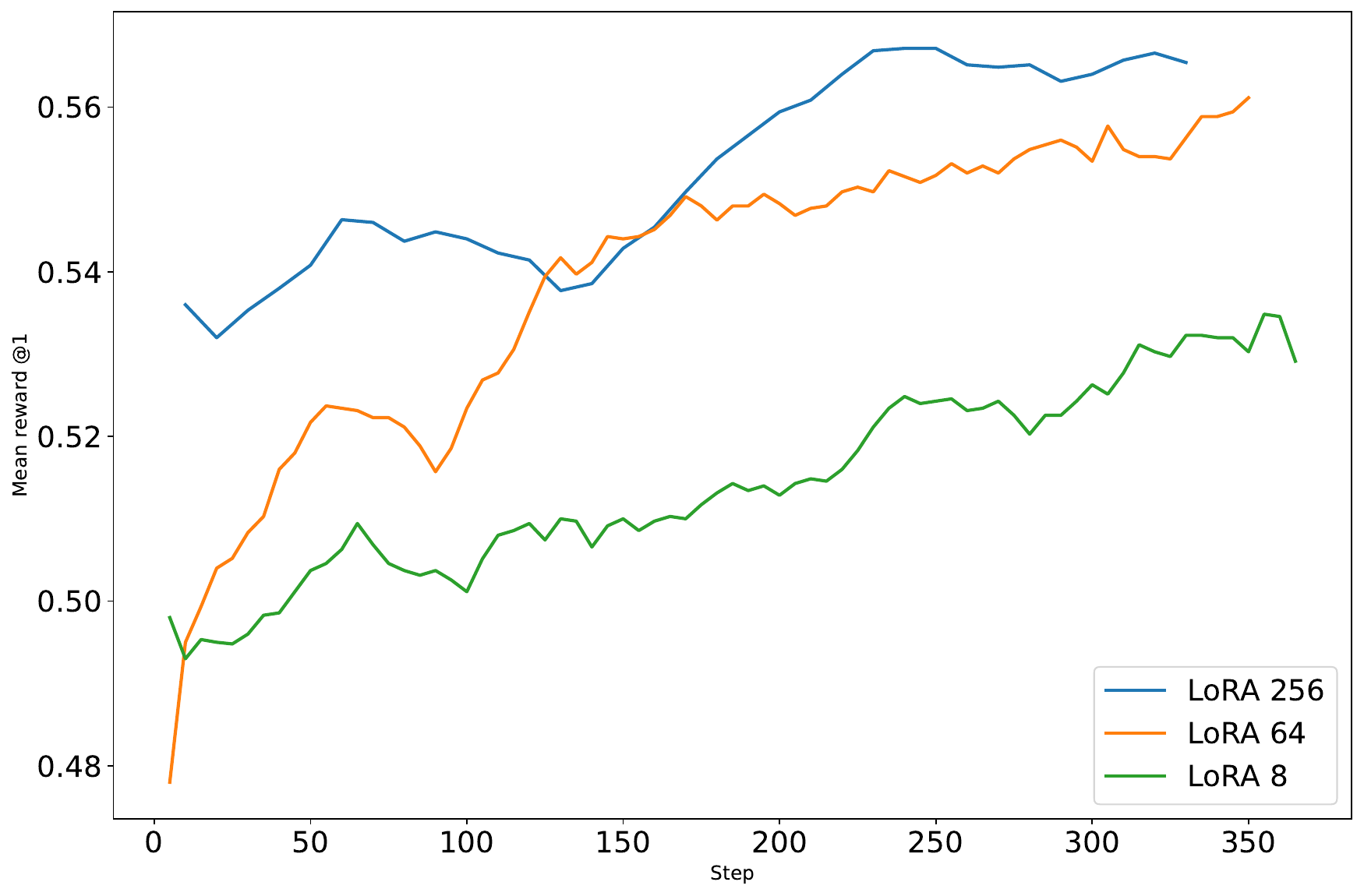}
        \caption{Qwen2.5-1.5B-Instruct}
    \end{subfigure}
    \hfill
    \begin{subfigure}{0.32\textwidth}
        \includegraphics[width=\linewidth]{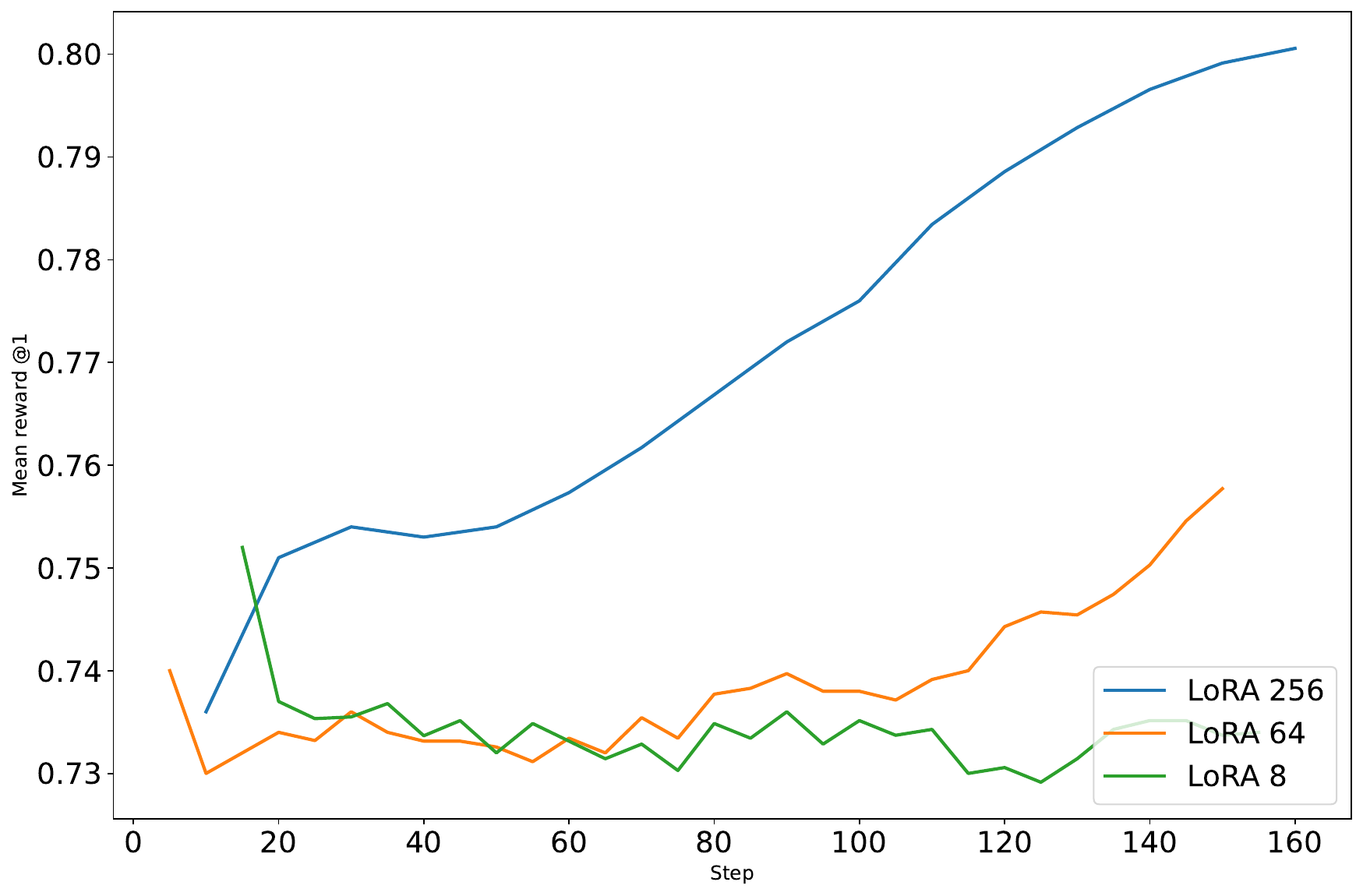}
        \caption{DeepScaleR-1.5B}
    \end{subfigure}
    \hfill
    \begin{subfigure}{0.32\textwidth}
        \includegraphics[width=\linewidth]{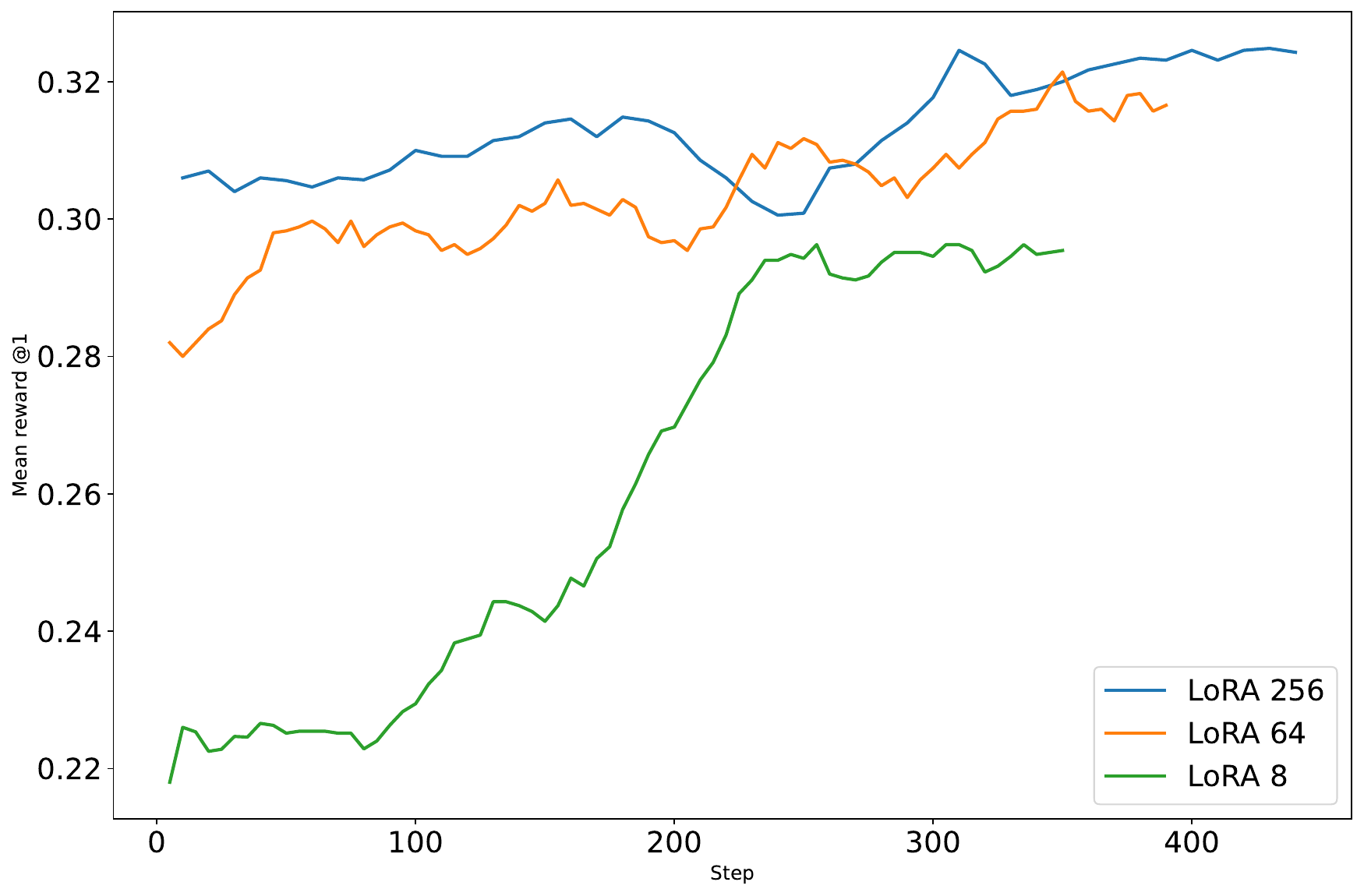}
        \caption{Llama-3.2-1B}
    \end{subfigure}    
    \vspace{0.5em}
    
    \begin{subfigure}{0.42\textwidth}
        \includegraphics[width=\linewidth]{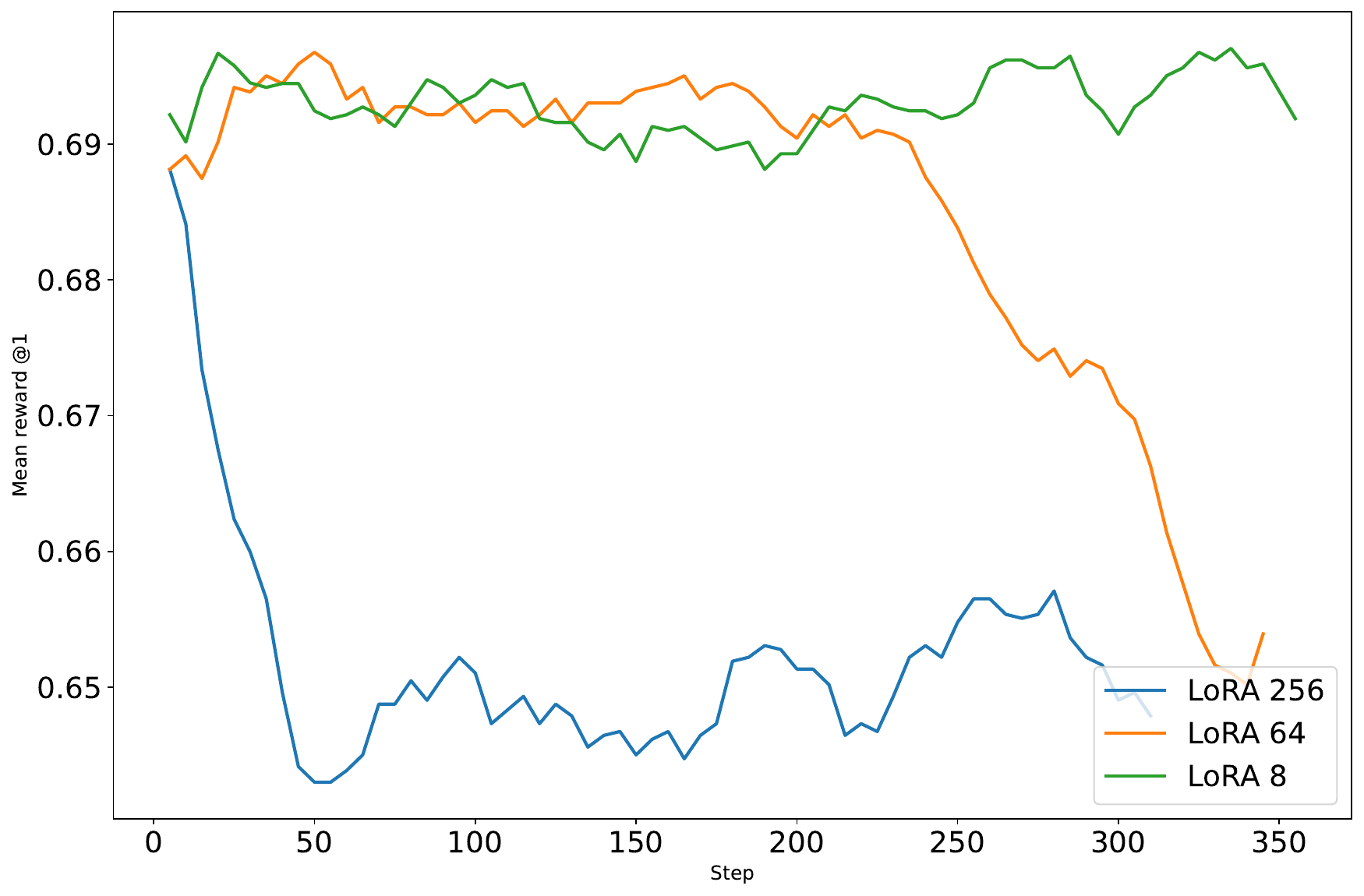}
        \caption{Qwen2.5-Math-1.5B}
    \end{subfigure}
    \hfill
    \begin{subfigure}{0.42\textwidth}
        \includegraphics[width=\linewidth]{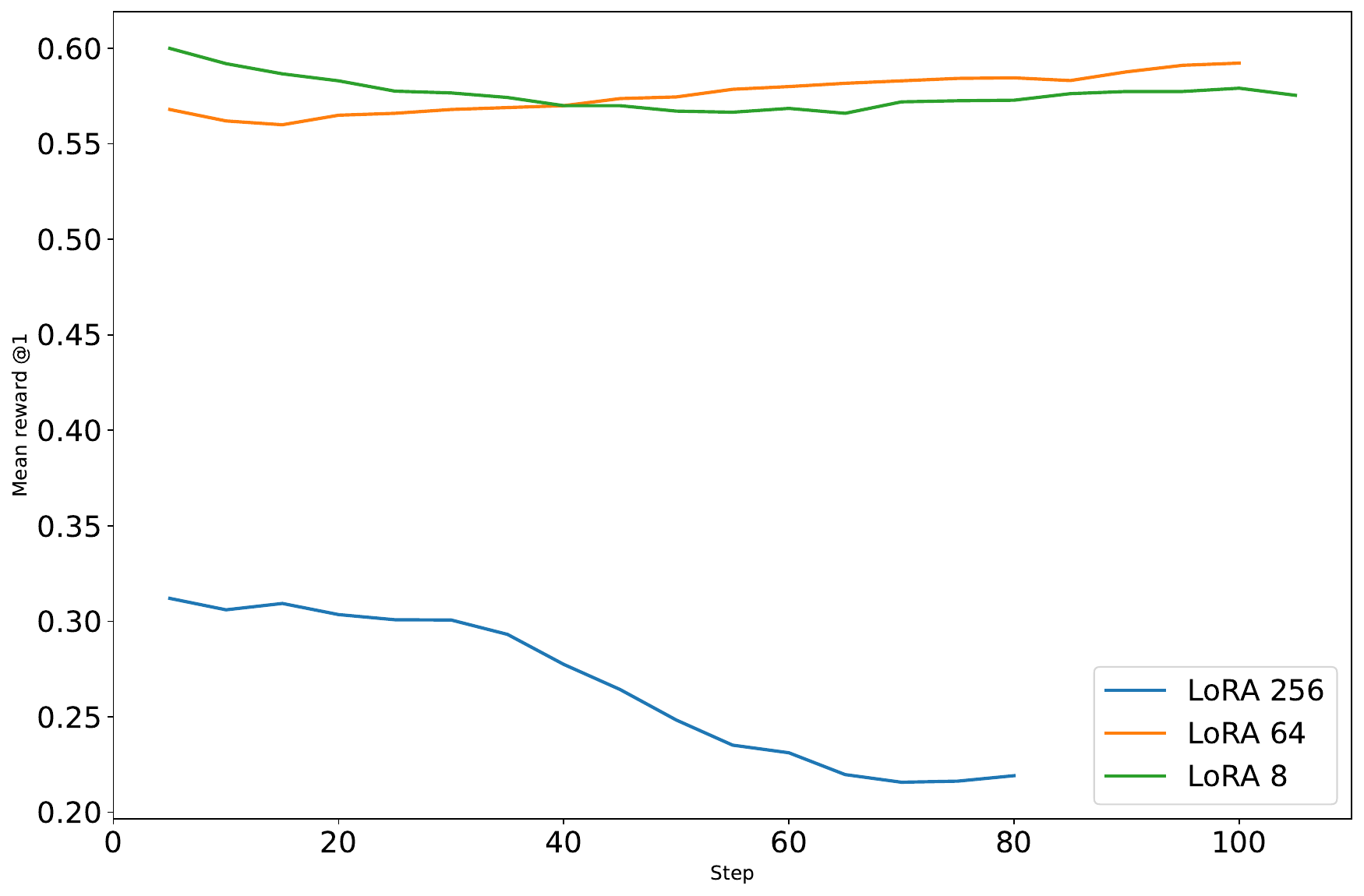}
        \caption{Qwen3-0.6B}
    \end{subfigure}
    \caption{\textbf{Validation Accuracy on MATH500.} Successful models (Top Row) show correlation between training reward and validation score. Qwen-Math (Bottom Center) exhibits "specialist collapse" at high ranks.}
    \label{fig:math500}
\end{figure*}

To ensure the reward optimization translates to actual reasoning capability, we tracked Zero-Shot Pass@1 on the MATH500 benchmark throughout training.
Figure \ref{fig:math500} confirms the "damage vs. help" trade-off:
\begin{itemize}
    \item \textbf{The Learners:} DeepScaleR-1.5B ($r=256$) and Qwen2.5-1.5B-Instruct ($r=256$) show strong, consistent gains in validation accuracy. The gains in the former model are much higher than that of the latter, and we attribute this to the former adjusting moreso to the reward function as compared to learning new reasoning abilities, which is likely happening in the latter.
    \item \textbf{The Collapse:} Qwen2.5-Math-1.5B suffers a sharp performance crash at Rank 256. The RL updates actively harmed its ability to solve math problems compared to its initialization. Another similar pattern is observed with Qwen3-0.6B. It is important to note here that the higher rank updates may lead to a collapse, but the lower rank updates mostly allow the model to stay stagnant.
\end{itemize}

\subsection{Evolution of Response Length}
\begin{figure*}[t]
    \centering
    \begin{subfigure}{0.48\textwidth}
        \includegraphics[width=\linewidth]{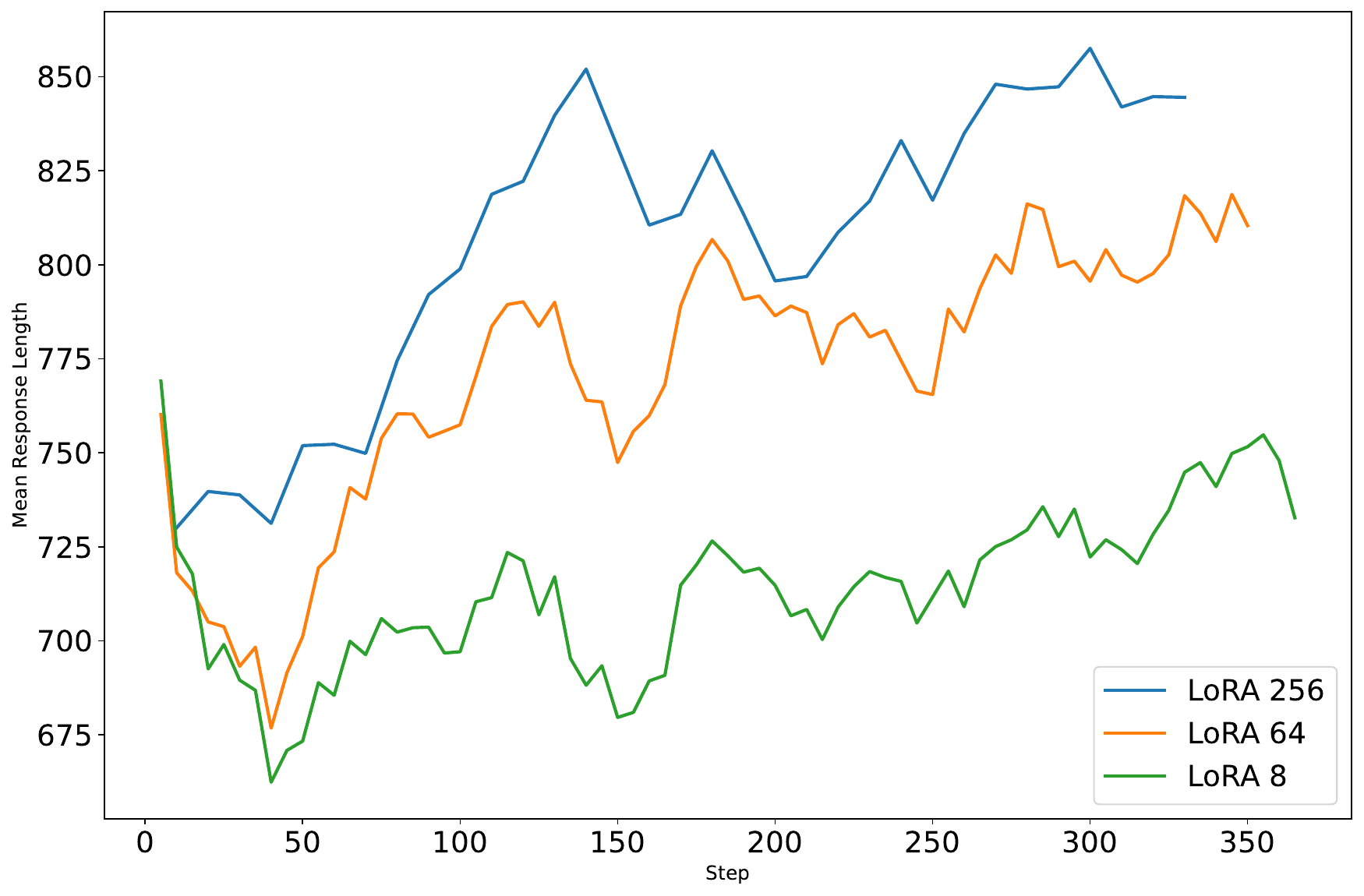}
        \caption{Qwen2.5-1.5B-Instruct}
    \end{subfigure}
    \hfill
    \begin{subfigure}{0.48\textwidth}
        \includegraphics[width=\linewidth]{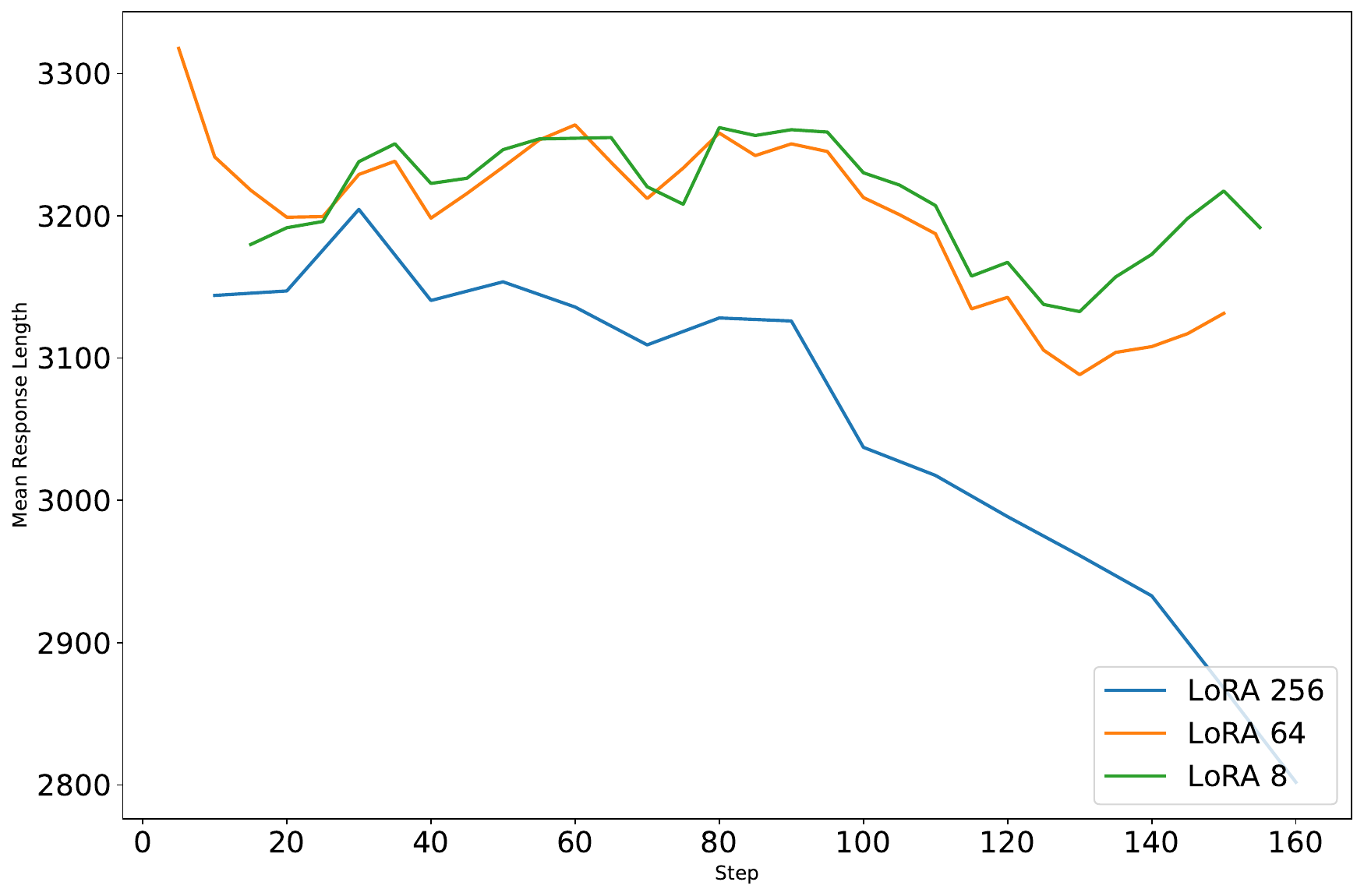}
        \caption{DeepScaleR-1.5B}
    \end{subfigure}
    
    \vspace{0.5em}
    
    \begin{subfigure}{0.32\textwidth}
        \includegraphics[width=\linewidth]{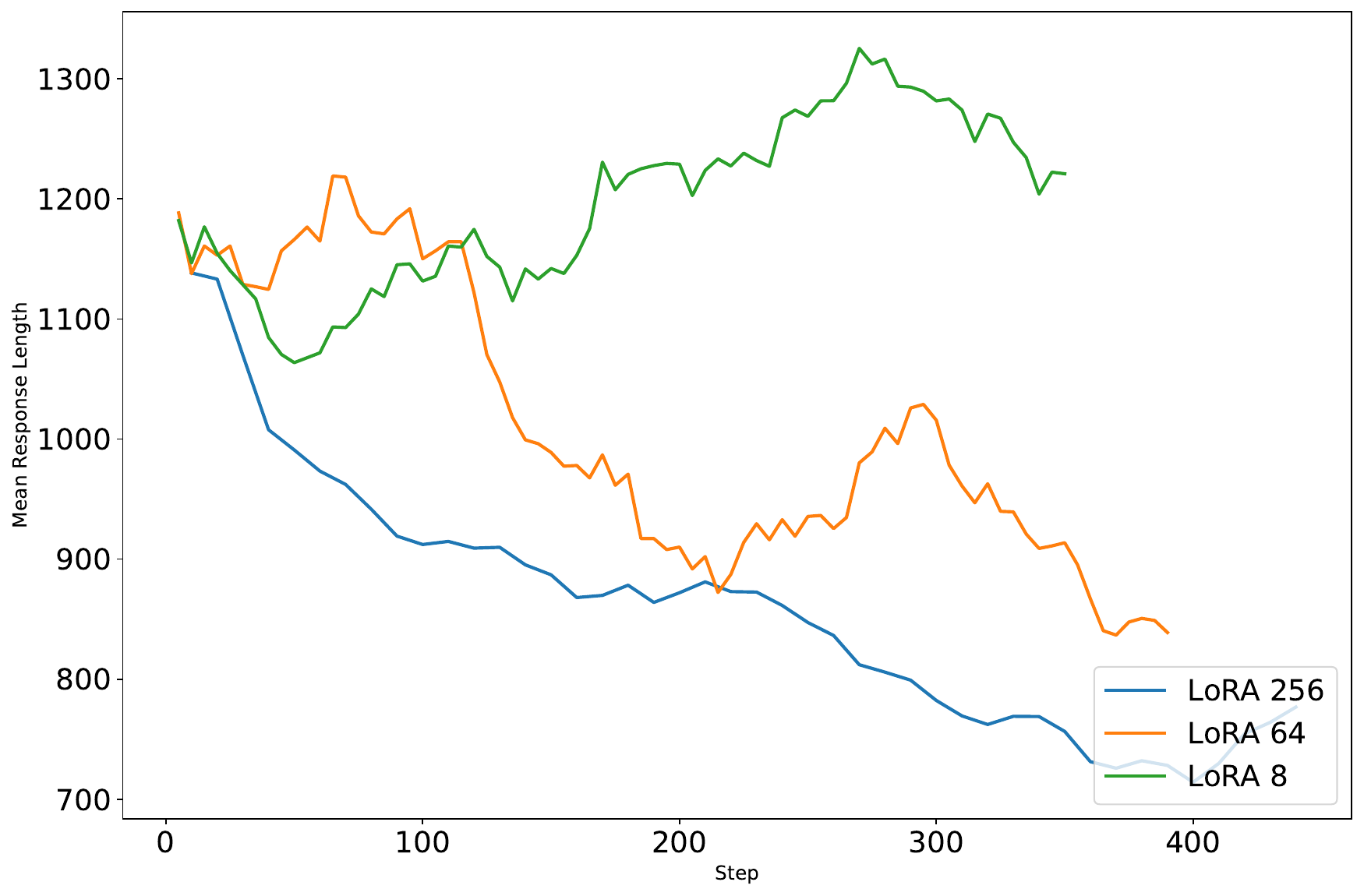}
        \caption{Llama-3.2-1B}
    \end{subfigure}
    \hfill
    \begin{subfigure}{0.32\textwidth}
        \includegraphics[width=\linewidth]{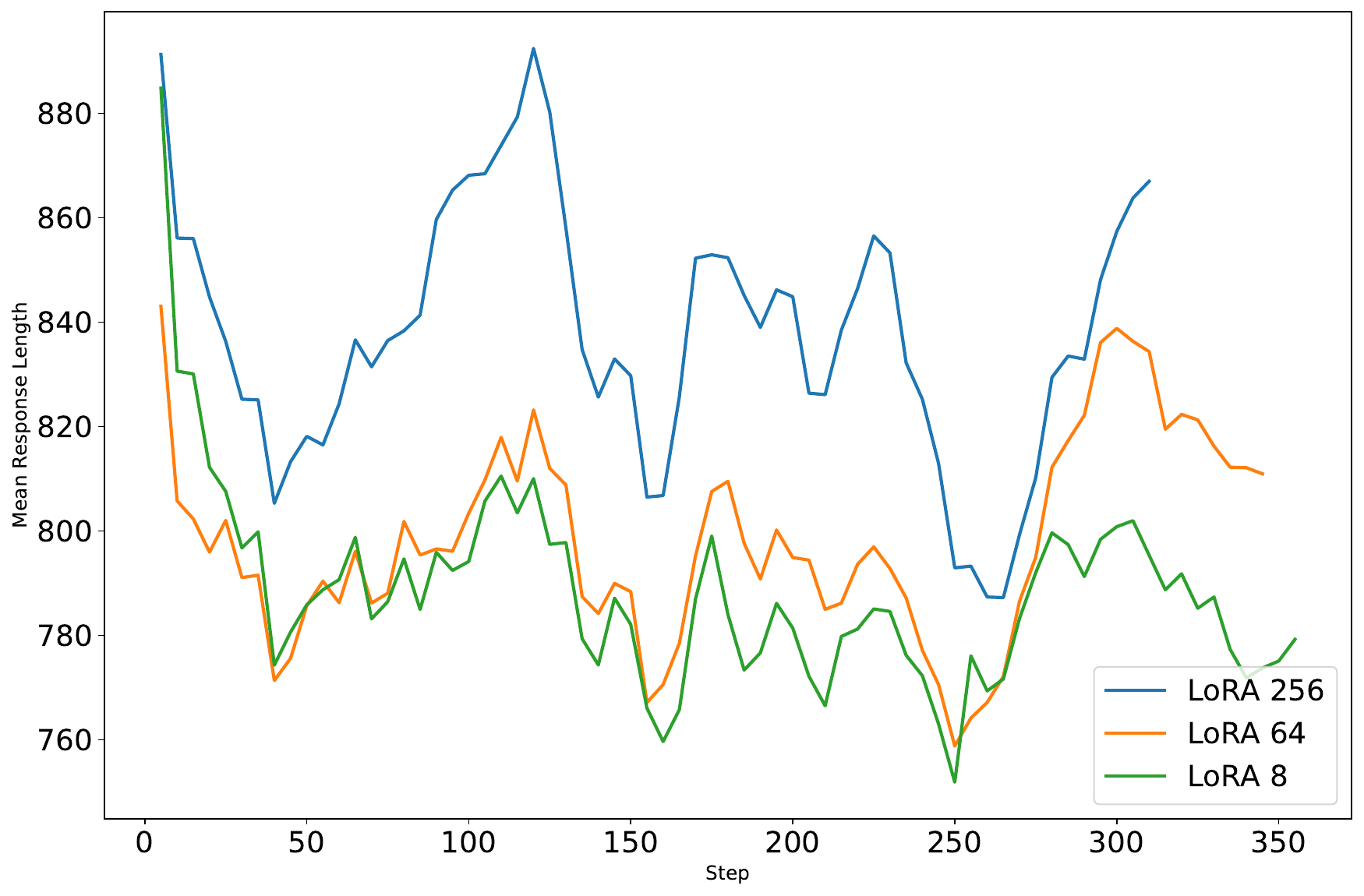}
        \caption{Qwen2.5-Math-1.5B}
    \end{subfigure}
    \hfill
    \begin{subfigure}{0.32\textwidth}
        \includegraphics[width=\linewidth]{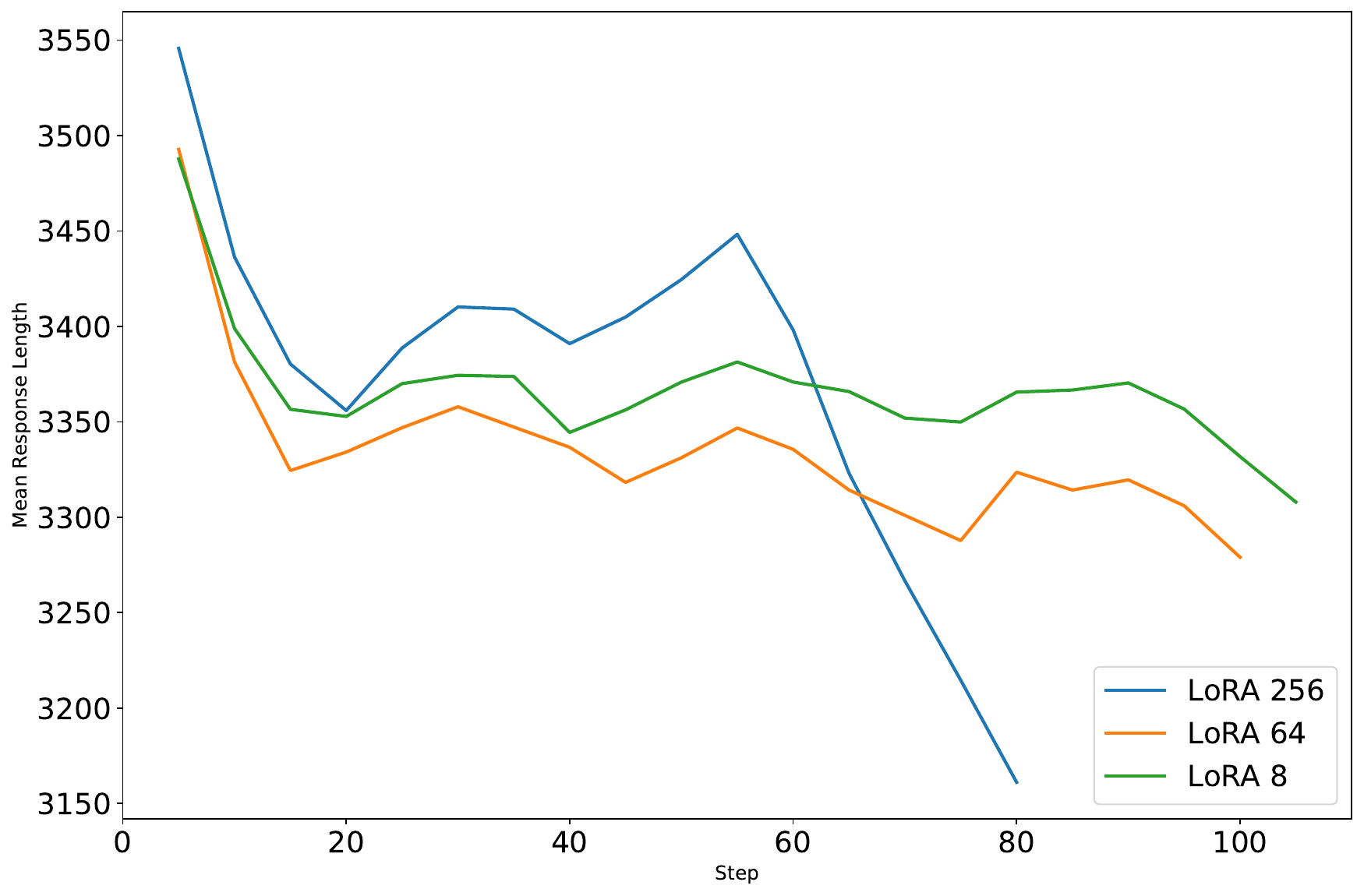}
        \caption{Qwen3-0.6B}
    \end{subfigure}
    \caption{\textbf{Evolution of Response Length.} Plastic models (Top Row) dynamically increased their context usage (``thinking") to maximize reward. Rigid models (Bottom Row) failed to adapt or suffered length collapse.}
    \label{fig:length}
\end{figure*}

We analyzed the average response length (number of generated tokens) to understand the mechanism behind the performance gains.
As seen in Figure \ref{fig:length}, plasticity relates well with actual test-time-compute albeit not having a consistent pattern:
\begin{itemize}
    \item \textbf{Expansion:} Llama-3.2-1B and Qwen2.5-1.5B-Instruct demonstrated active exploration by elongating their reasoning chains. Notably, Llama-3.2-1B nearly doubled its response length from $\sim$700 to over 1,200 tokens. DeepScaleR, while already starting with a long context ($\sim$3,150 tokens) may have learned conciseness in reasoning owing to the limited token budget compared to its previous post-training runs.
    \item \textbf{Contraction:} In contrast, the failing or saturated models (Qwen3-0.6B and Qwen2.5-Math-1.5B) reverted to shorter or unstable responses. Qwen3-0.6B, for instance, saw its response length contract, correlating with its inability to improve validation performance.
\end{itemize}

\subsection{Evaluation}

To assess whether the training gains observed on MATH500 translate to robust generalization, we evaluated the final checkpoints on three held-out competition benchmarks: \textbf{AMC 23}, \textbf{AIME 24}, and \textbf{AIME 25}. We report Zero-Shot Pass@1, as well as Pass@8 and Pass@16 to gauge the models' consistency.

\paragraph{Benchmark Performance.}
As detailed in Table \ref{tab:combined_results}, the impact of low-budget RLVR varies dramatically across model families:

\begin{itemize}
    \item \textbf{DeepScaleR-1.5B} shows the greatest improvement over all benchmarks in all Pass@k estimates. It showed significant gains over its baseline and serves as evidence that LoRA finetuning for a capable instruction-finetuned base can be very effective.
    
    \item \textbf{Qwen2.5-Math-1.5B and Qwen3-0.6B} show improvements in some benchmarks, albeit there is no consistent pattern. Interestingly we can note how the Pass@8 and Pass@16 are more likely to improve than Pass@1, pointing to how the latent reasoning abilities are still improving. We can also recall that the validation scores on MATH500 collapsed for higher ranks, but mostly stayed stagnant for lower ranks which is reflected in these measures. This backs up one of our hypotheses surrounding cheap noisy RL updates disrupting the fragile manifold of heavily pre-optimized models (be it for solving math problems or reasoning, respectively). They would likely require many more training steps.

    \item \textbf{Qwen2.5-1.5B-Instruct and Llama-3.2-1B} did not really show any changes from the baseline when it came to these much harder problems \textit{even though} we saw decent gains in MATH500 scores. Unlike the aforementioned collapse, these models showed minor fluctuations or stagnancy, suggesting that while they possess plasticity, they may require a longer "warm-up" period or more data than the 24-hour budget allowed to bridge the gap to expert reasoning. This may stem from how a benchmark like MATH500 is much easier than soemthing like AIME24 and hence reflects marginal improvements in reasoning ability better.
\end{itemize}

\begin{table*}[h]
    \centering
    \caption{Comparison of Baseline vs. Final (LoRA) performance. \textbf{Bold} values indicate improvement over the baseline.}
    \label{tab:combined_results}
        \begin{tabular}{l l | ccc | ccc | ccc}
            \toprule
            \multirow{2}{*}{\textbf{Model}} & \multirow{2}{*}{\textbf{Config}} & \multicolumn{3}{c}{\textbf{AIME 24 (\%)}} & \multicolumn{3}{c}{\textbf{AIME 25 (\%)}} & \multicolumn{3}{c}{\textbf{AMC 23 (\%)}} \\
             & & @1 & @8 & @16 & @1 & @8 & @16 & @1 & @8 & @16 \\
            \midrule
            
            \multirow{2}{*}{DeepScaleR-1.5B-Preview} 
                & Baseline & 28.9 & 53.6 & 62.5 & 17.7 & 35.8 & 45.2 & 58.8 & 87.5 & 92.2 \\
                & Final & \textbf{40.0} & \textbf{68.0} & \textbf{70.0} & \textbf{28.1} & \textbf{50.2} & \textbf{56.7} & \textbf{79.2} & \textbf{96.0} & \textbf{97.5} \\
            \midrule

            \multirow{2}{*}{Qwen2.5-Instruct-1.5B} 
                & Baseline & 2.5 & 10.0 & 16.7 & 0.6 & 4.2 & 6.7 & 24.2 & 57.7 & 65.0 \\
                & Final & 2.5 & 10.0 & 16.6 & 0.6 & 4.2 & 6.7 & 24.2 & 57.6 & 65.0 \\
            \midrule

            \multirow{2}{*}{Llama-3.2-1B} 
                & Baseline & 1.0 & 7.6 & 13.3 & 0.2 & 1.7 & 3.3 & 10.6 & 34.8 & 50.0 \\
                & Final & 1.0 & 7.6 & 13.3 & 0.2 & 1.7 & 3.3 & 10.6 & 34.7 & 50.0 \\
            \midrule

            \multirow{2}{*}{Qwen2.5-Math-1.5B} 
                & Baseline & 3.8 & 10.6 & 13.3 & 2.5 & 14.0 & 20.0 & 22.2 & 59.7 & 72.5 \\
                & Final & \textbf{4.8} & \textbf{15.1} & \textbf{20.0} & \textbf{2.9} & \textbf{14.4} & 20.0 & 18.9 & \textbf{62.1} & \textbf{77.5} \\
            \midrule

            \multirow{2}{*}{Qwen3-0.6B} 
                & Baseline & 9.4 & 28.4 & 36.7 & 14.0 & 31.3 & 36.7 & 46.9 & 78.2 & 85.0 \\
                & Final & 8.5 & \textbf{28.5} & 36.7 & \textbf{14.6} & \textbf{31.9} & 36.7 & \textbf{48.3} & 77.0 & 82.5 \\
            
            \bottomrule
        \end{tabular}
\end{table*}

\subsection{Entropy Dynamics and Policy Divergence}

\begin{table}[!ht]
    \centering
    \caption{Relative Change in Policy Entropy ($\%$) by Rank. {High-rank adapters ($r=256$) drive massive entropy shifts compared to $r=8$.}}
    \label{tab:entropy_relative_change}
    
    \resizebox{\linewidth}{!}{
        \begin{tabular}{l ccc}
            \toprule
            \multirow{2}{*}{\textbf{Model}} & \multicolumn{3}{c}{\textbf{Relative Entropy Change ($\Delta H_{\text{rel}}$)}} \\
            \cmidrule(l){2-4}
             & \textbf{Rank 8} & \textbf{Rank 64} & \textbf{Rank 256} \\
            \midrule
            DeepScaleR-1.5B & -6.7 & -2.7 & \textbf{-27.0} \\
            Llama-3.2-1B & -13.7 & -67.9 & \textbf{-92.5} \\
            Qwen2.5-1.5B-Instruct & -5.5 & -62.1 & \textbf{-60.1} \\
            Qwen2.5-Math-1.5B & +37.7 & -3.8 & \textbf{-31.7} \\
            Qwen3-0.6B & -2.0 & -11.0 & \textbf{-61.5} \\
            \bottomrule
        \end{tabular}
    }
\end{table}

To understand the mechanism of adaptation under strict compute constraints, we additionally analyze the evolution of the model's policy entropy throughout training. We define the mean token-level entropy $H(\pi)$ for a response sequence $y$ given prompt $x$ as:
\begin{equation}
    H(\pi) = -\frac{1}{T} \sum_{t=1}^{T} \sum_{v \in V} \pi(v | x, y_{<t}) \log \pi(v | x, y_{<t})
\end{equation}
Recent theoretical work suggests that reinforcement learning acts as an entropy regulation mechanism, where the model trades policy entropy (uncertainty) for higher expected reward \cite{cui2025entropymechanismreinforcementlearning}. We track the relative change in this metric to quantify how far the fine-tuned policy diverges from its initialization, as show in Table~\ref{tab:entropy_relative_change}.

\paragraph{Rank-Dependent Capacity.}
We observe that the magnitude of policy divergence is heavily influenced by adapter rank. Across all architectures, models trained with rank $r=256$ exhibited relative entropy shifts up to 3x larger than those with $r=8$. This confirms that low-rank constraints mechanically limit the policy's ability to deviate from the pre-trained manifold, effectively anchoring the model to its initialization regardless of the gradient signal.

\paragraph{Divergence vs. Directed Exploration.}
However, high policy divergence is a necessary but insufficient condition for performance capability. Both DeepScaleR-1.5B and Llama-3.2-1B exhibited significant entropy volatility at high ranks, yet their outcomes diverged. DeepScaleR-1.5B utilized this capacity to explore valid reasoning paths (increasing Pass@16), whereas Llama-3.2-1B, lacking strong reasoning priors, drifted stochastically without converging on high-reward regions.

\paragraph{Optimization Collapse in Aligned Models.}
For models that are already heavily optimized for the target task (e.g., Qwen2.5-Math-1.5B), high-rank updates acted as destructive interference. Instead of refining the policy, the noisy RL gradients caused a sharp reduction in entropy (mode collapse). The model effectively retreated to low-entropy, safety-seeking behaviors (such as short responses) rather than exploring the solution space, leading to performance degradation.

\section{Discussion \& Future Work}

\paragraph{The Latent Reasoning Gap \& Entropy Dissociation.}
We find that the delta between Pass@1 and Pass@16 acts as a critical feasibility signal for RLVR. A large gap (e.g., {DeepScaleR-1.5B-Preview}'s 40\% vs. 70\%) indicates "latent" capability that GRPO can effectively bootstrap. This observation complicates recent findings on the ``Entropy Mechanism" \cite{cui2025entropymechanismreinforcementlearning}, which posit that performance improvements strictly trade off with policy entropy. While valid for capable models, our results with {Llama-3.2-1B-Instruct} challenge this universality: the model exhibited significant entropy reduction (collapse) without corresponding performance gains. This suggests that for models with weak reasoning priors, entropy reduction may  signal a regression into simple convergent behaviors rather than optimization, dissociating the link between certainty and correctness as observed in contemporary works on stronger reasoning models.

\paragraph{The ``Warm-Start" Hypothesis.}
Our results reinforce the ``LIMA hypothesis" \cite{zhou2023limaalignment} within an RL context: RLVR acts primarily as an alignment mechanism to expose latent knowledge, rather than a pedagogical tool to teach new theorems. {Llama-3.2-1B-Instruct}'s failure suggests a ``Cold Start" problem where random exploration cannot bridge the gap to the first non-zero reward. We posit that RLVR is most cost-effective when applied to "warm" models—those already seeded with reasoning behaviors via pre-training or SFT—allowing the optimizer to focus on utilizing the latent space rather than constructing it. Future work should investigate brief SFT phases as a warm-up for reasoning``Reasoning Warm-up" to prime off-the-shelf models before RLVR.

\paragraph{Algorithmic Constraints \& Policy Deviations.}
The "rigidity" observed in {Qwen2.5-Math-1.5B} suggests that the conservative trust-region constraints of standard PPO/GRPO may be counter-productive when fine-tuning specialists with noisy, micro-budget updates. We hypothesize that algorithms which relax the aggressive clipping of the policy gradient—such as Dr. GRPO \cite{liu2025understandingr1zeroliketrainingcritical} or DAPO/CLIP-Higher \cite{yu2025dapoopensourcellmreinforcement}—could allow the policy to deviate sufficiently from its local optimum to discover more robust reasoning paths. By permitting larger distinct updates, these methods might prevent the mode collapse we observed in specialists, provided the reward signal remains verifiable.

\paragraph{Scaling Laws of High-Rank Adaptation.}
Finally, our success with high-rank LoRA ($r=256$) on {DeepScaleR-1.5B-Preview} suggests a scalable paradigm for reasoning alignment: treating high-rank adapters as a cost-effective alternative to full-parameter fine-tuning. If RLVR is largely about surface-level alignment of latent reasoning (as seen in DeepSeek-R1), then massive full-parameter updates may be redundant. Future work should extend this study to larger scales, comparing high-rank LoRA against full-finetuning over extended epochs to determine if the ``plasticity" provided by $r=256$ is sufficient to replicate the gains of full-scale training at a fraction of the GPU-hour cost.

\section{Conclusion}
Our investigation into reasoning alignment under strict compute constraints reveals that high-performance mathematical reasoning is attainable on a ``micro-budget", provided the alignment strategy matches the model's initialization. We demonstrate that plasticity is the governing resource: generalist models like {Qwen2.5-1.5B-Instruct} and {DeepScaleR-1.5B} possess the latent capacity to actively explore and internalize reasoning behaviors when empowered by high-rank adapters ($r=256$), enabling {DeepScaleR} to achieve a state-of-the-art {40.0\% Pass@1 on AIME 24}. Conversely, the rigidity of heavily optimized models like {Qwen2.5-Math} renders them vulnerable to the noisy updates of low-budget RL, leading to performance collapse. Ultimately, we propose that the most efficient path to democratizing reasoning is not to incrementally refine experts, but to catalyze generalists—using high-rank adaptation to unlock the latent reasoning capabilities already present in their pre-trained manifolds.

\section*{Limitations}

Our study was designed to probe the feasibility of reasoning alignment under extreme constraints, and as such, several limitations apply to our findings:

\begin{itemize}
    \item \textbf{Model Scale:} Due to the single-GPU memory constraint (48GB), our investigation was restricted to small language models ($\leq1.5\text{B}$ parameters). It remains verifying whether the ``plasticity vs. rigidity" trade-off we observed holds for larger architectures (e.g. 7B,8B,32B models), which often possess more robust internal representations and might be more resilient to noisy LoRA updates.
    
    \item \textbf{Hyperparameter Scope:} The strict 24-hour compute budget precluded a comprehensive grid search. We utilized fixed values for critical hyperparameters such as learning rate and LoRA alpha across all runs. It is possible that the ``collapse" observed in some models could be mitigated with a more conservative learning rate or a tuned alpha/rank ratio, rather than being an intrinsic failure of the method itself.
    
    \item \textbf{Training Duration:} We limited training to a maximum of 24 hours ($\sim$300 update steps). While sufficient to observe divergence in plasticity, this window may be too short for "slow-learning" generalist models to fully converge. Longer training horizons might reveal that models like {Llama-3.2-1B} eventually overcome the "cold start" problem given enough exploration time.
    
    \item \textbf{Single-Seed Stochasticity:} Finally, due to resource limitations, each experimental configuration was conducted with a single random seed. Given the inherent high variance of reinforcement learning—particularly with the GRPO estimator—our results may be influenced by initialization noise. Future work with greater resources should employ multi-seed averaging to report confidence intervals and ensure statistical significance.
\end{itemize}

\bibliography{custom}

@misc{zhou2023limaalignment,
      title={LIMA: Less Is More for Alignment}, 
      author={Chunting Zhou and Pengfei Liu and Puxin Xu and Srini Iyer and Jiao Sun and Yuning Mao and Xuezhe Ma and Avia Efrat and Ping Yu and Lili Yu and Susan Zhang and Gargi Ghosh and Mike Lewis and Luke Zettlemoyer and Omer Levy},
      year={2023},
      eprint={2305.11206},
      archivePrefix={arXiv},
      primaryClass={cs.CL},
      url={https://arxiv.org/abs/2305.11206}, 
}

@misc{yu2025dapoopensourcellmreinforcement,
      title={DAPO: An Open-Source LLM Reinforcement Learning System at Scale}, 
      author={Qiying Yu and Zheng Zhang and Ruofei Zhu and Yufeng Yuan and Xiaochen Zuo and Yu Yue and Weinan Dai and Tiantian Fan and Gaohong Liu and Lingjun Liu and Xin Liu and Haibin Lin and Zhiqi Lin and Bole Ma and Guangming Sheng and Yuxuan Tong and Chi Zhang and Mofan Zhang and Wang Zhang and Hang Zhu and Jinhua Zhu and Jiaze Chen and Jiangjie Chen and Chengyi Wang and Hongli Yu and Yuxuan Song and Xiangpeng Wei and Hao Zhou and Jingjing Liu and Wei-Ying Ma and Ya-Qin Zhang and Lin Yan and Mu Qiao and Yonghui Wu and Mingxuan Wang},
      year={2025},
      eprint={2503.14476},
      archivePrefix={arXiv},
      primaryClass={cs.LG},
      url={https://arxiv.org/abs/2503.14476}, 
}

@misc{cui2025entropymechanismreinforcementlearning,
      title={The Entropy Mechanism of Reinforcement Learning for Reasoning Language Models}, 
      author={Ganqu Cui and Yuchen Zhang and Jiacheng Chen and Lifan Yuan and Zhi Wang and Yuxin Zuo and Haozhan Li and Yuchen Fan and Huayu Chen and Weize Chen and Zhiyuan Liu and Hao Peng and Lei Bai and Wanli Ouyang and Yu Cheng and Bowen Zhou and Ning Ding},
      year={2025},
      eprint={2505.22617},
      archivePrefix={arXiv},
      primaryClass={cs.LG},
      url={https://arxiv.org/abs/2505.22617}, 
}

@misc{deepscaler2025,
  title={DeepScaleR: Surpassing O1-Preview with a 1.5B Model by Scaling RL},
  author={Michael Luo and Sijun Tan and Justin Wong and Xiaoxiang Shi and William Y. Tang and Manan Roongta and Colin Cai and Jeffrey Luo and Li Erran Li and Raluca Ada Popa and Ion Stoica},
  howpublished={\url{https://pretty-radio-b75.notion.site/DeepScaleR-Surpassing-O1-Preview-with-a-1-5B-Model-by-Scaling-RL-19681902c1468005bed8ca303013a4e2}},
  note={Notion Blog},
  year={2025}
}

@misc{dang2025reinforcementlearningreasoningsmall,
      title={Reinforcement Learning for Reasoning in Small LLMs: What Works and What Doesn't}, 
      author={Quy-Anh Dang and Chris Ngo},
      year={2025},
      eprint={2503.16219},
      archivePrefix={arXiv},
      primaryClass={cs.LG},
      url={https://arxiv.org/abs/2503.16219}, 
}

@misc{liu2025understandingr1zeroliketrainingcritical,
      title={Understanding R1-Zero-Like Training: A Critical Perspective}, 
      author={Zichen Liu and Changyu Chen and Wenjun Li and Penghui Qi and Tianyu Pang and Chao Du and Wee Sun Lee and Min Lin},
      year={2025},
      eprint={2503.20783},
      archivePrefix={arXiv},
      primaryClass={cs.LG},
      url={https://arxiv.org/abs/2503.20783}, 
}

@misc{muennighoff2025s1simpletesttimescaling,
      title={s1: Simple test-time scaling}, 
      author={Niklas Muennighoff and Zitong Yang and Weijia Shi and Xiang Lisa Li and Li Fei-Fei and Hannaneh Hajishirzi and Luke Zettlemoyer and Percy Liang and Emmanuel Candès and Tatsunori Hashimoto},
      year={2025},
      eprint={2501.19393},
      archivePrefix={arXiv},
      primaryClass={cs.CL},
      url={https://arxiv.org/abs/2501.19393}, 
}

@misc{ye2025limoreasoning,
      title={LIMO: Less is More for Reasoning}, 
      author={Yixin Ye and Zhen Huang and Yang Xiao and Ethan Chern and Shijie Xia and Pengfei Liu},
      year={2025},
      eprint={2502.03387},
      archivePrefix={arXiv},
      primaryClass={cs.CL},
      url={https://arxiv.org/abs/2502.03387}, 
}

@article{schulman2025lora,
  author = {John Schulman and Thinking Machines Lab},
  title = {LoRA Without Regret},
  journal = {Thinking Machines Lab: Connectionism},
  year = {2025},
  note = {https://thinkingmachines.ai/blog/lora/},
  doi = {10.64434/tml.20250929},
}

@misc{wang2025tinatinyreasoningmodels,
      title={Tina: Tiny Reasoning Models via LoRA}, 
      author={Shangshang Wang and Julian Asilis and Ömer Faruk Akgül and Enes Burak Bilgin and Ollie Liu and Willie Neiswanger},
      year={2025},
      eprint={2504.15777},
      archivePrefix={arXiv},
      primaryClass={cs.CL},
      url={https://arxiv.org/abs/2504.15777}, 
}

@misc{shao2024deepseekmathpushinglimitsmathematical,
      title={DeepSeekMath: Pushing the Limits of Mathematical Reasoning in Open Language Models}, 
      author={Zhihong Shao and Peiyi Wang and Qihao Zhu and Runxin Xu and Junxiao Song and Xiao Bi and Haowei Zhang and Mingchuan Zhang and Y. K. Li and Y. Wu and Daya Guo},
      year={2024},
      eprint={2402.03300},
      archivePrefix={arXiv},
      primaryClass={cs.CL},
      url={https://arxiv.org/abs/2402.03300}, 
}

@misc{wei2023chainofthoughtpromptingelicitsreasoning,
      title={Chain-of-Thought Prompting Elicits Reasoning in Large Language Models}, 
      author={Jason Wei and Xuezhi Wang and Dale Schuurmans and Maarten Bosma and Brian Ichter and Fei Xia and Ed Chi and Quoc Le and Denny Zhou},
      year={2023},
      eprint={2201.11903},
      archivePrefix={arXiv},
      primaryClass={cs.CL},
      url={https://arxiv.org/abs/2201.11903}, 
}

@misc{openai2024gpt4technicalreport,
      title={GPT-4 Technical Report}, 
      author={OpenAI and Josh Achiam and Steven Adler and Sandhini Agarwal and Lama Ahmad and Ilge Akkaya and Florencia Leoni Aleman and Diogo Almeida and Janko Altenschmidt and Sam Altman and Shyamal Anadkat and Red Avila and Igor Babuschkin and Suchir Balaji and Valerie Balcom and Paul Baltescu and Haiming Bao and Mohammad Bavarian and Jeff Belgum and Irwan Bello and Jake Berdine and Gabriel Bernadett-Shapiro and Christopher Berner and Lenny Bogdonoff and Oleg Boiko and Madelaine Boyd and Anna-Luisa Brakman and Greg Brockman and Tim Brooks and Miles Brundage and Kevin Button and Trevor Cai and Rosie Campbell and Andrew Cann and Brittany Carey and Chelsea Carlson and Rory Carmichael and Brooke Chan and Che Chang and Fotis Chantzis and Derek Chen and Sully Chen and Ruby Chen and Jason Chen and Mark Chen and Ben Chess and Chester Cho and Casey Chu and Hyung Won Chung and Dave Cummings and Jeremiah Currier and Yunxing Dai and Cory Decareaux and Thomas Degry and Noah Deutsch and Damien Deville and Arka Dhar and David Dohan and Steve Dowling and Sheila Dunning and Adrien Ecoffet and Atty Eleti and Tyna Eloundou and David Farhi and Liam Fedus and Niko Felix and Simón Posada Fishman and Juston Forte and Isabella Fulford and Leo Gao and Elie Georges and Christian Gibson and Vik Goel and Tarun Gogineni and Gabriel Goh and Rapha Gontijo-Lopes and Jonathan Gordon and Morgan Grafstein and Scott Gray and Ryan Greene and Joshua Gross and Shixiang Shane Gu and Yufei Guo and Chris Hallacy and Jesse Han and Jeff Harris and Yuchen He and Mike Heaton and Johannes Heidecke and Chris Hesse and Alan Hickey and Wade Hickey and Peter Hoeschele and Brandon Houghton and Kenny Hsu and Shengli Hu and Xin Hu and Joost Huizinga and Shantanu Jain and Shawn Jain and Joanne Jang and Angela Jiang and Roger Jiang and Haozhun Jin and Denny Jin and Shino Jomoto and Billie Jonn and Heewoo Jun and Tomer Kaftan and Łukasz Kaiser and Ali Kamali and Ingmar Kanitscheider and Nitish Shirish Keskar and Tabarak Khan and Logan Kilpatrick and Jong Wook Kim and Christina Kim and Yongjik Kim and Jan Hendrik Kirchner and Jamie Kiros and Matt Knight and Daniel Kokotajlo and Łukasz Kondraciuk and Andrew Kondrich and Aris Konstantinidis and Kyle Kosic and Gretchen Krueger and Vishal Kuo and Michael Lampe and Ikai Lan and Teddy Lee and Jan Leike and Jade Leung and Daniel Levy and Chak Ming Li and Rachel Lim and Molly Lin and Stephanie Lin and Mateusz Litwin and Theresa Lopez and Ryan Lowe and Patricia Lue and Anna Makanju and Kim Malfacini and Sam Manning and Todor Markov and Yaniv Markovski and Bianca Martin and Katie Mayer and Andrew Mayne and Bob McGrew and Scott Mayer McKinney and Christine McLeavey and Paul McMillan and Jake McNeil and David Medina and Aalok Mehta and Jacob Menick and Luke Metz and Andrey Mishchenko and Pamela Mishkin and Vinnie Monaco and Evan Morikawa and Daniel Mossing and Tong Mu and Mira Murati and Oleg Murk and David Mély and Ashvin Nair and Reiichiro Nakano and Rajeev Nayak and Arvind Neelakantan and Richard Ngo and Hyeonwoo Noh and Long Ouyang and Cullen O'Keefe and Jakub Pachocki and Alex Paino and Joe Palermo and Ashley Pantuliano and Giambattista Parascandolo and Joel Parish and Emy Parparita and Alex Passos and Mikhail Pavlov and Andrew Peng and Adam Perelman and Filipe de Avila Belbute Peres and Michael Petrov and Henrique Ponde de Oliveira Pinto and Michael and Pokorny and Michelle Pokrass and Vitchyr H. Pong and Tolly Powell and Alethea Power and Boris Power and Elizabeth Proehl and Raul Puri and Alec Radford and Jack Rae and Aditya Ramesh and Cameron Raymond and Francis Real and Kendra Rimbach and Carl Ross and Bob Rotsted and Henri Roussez and Nick Ryder and Mario Saltarelli and Ted Sanders and Shibani Santurkar and Girish Sastry and Heather Schmidt and David Schnurr and John Schulman and Daniel Selsam and Kyla Sheppard and Toki Sherbakov and Jessica Shieh and Sarah Shoker and Pranav Shyam and Szymon Sidor and Eric Sigler and Maddie Simens and Jordan Sitkin and Katarina Slama and Ian Sohl and Benjamin Sokolowsky and Yang Song and Natalie Staudacher and Felipe Petroski Such and Natalie Summers and Ilya Sutskever and Jie Tang and Nikolas Tezak and Madeleine B. Thompson and Phil Tillet and Amin Tootoonchian and Elizabeth Tseng and Preston Tuggle and Nick Turley and Jerry Tworek and Juan Felipe Cerón Uribe and Andrea Vallone and Arun Vijayvergiya and Chelsea Voss and Carroll Wainwright and Justin Jay Wang and Alvin Wang and Ben Wang and Jonathan Ward and Jason Wei and CJ Weinmann and Akila Welihinda and Peter Welinder and Jiayi Weng and Lilian Weng and Matt Wiethoff and Dave Willner and Clemens Winter and Samuel Wolrich and Hannah Wong and Lauren Workman and Sherwin Wu and Jeff Wu and Michael Wu and Kai Xiao and Tao Xu and Sarah Yoo and Kevin Yu and Qiming Yuan and Wojciech Zaremba and Rowan Zellers and Chong Zhang and Marvin Zhang and Shengjia Zhao and Tianhao Zheng and Juntang Zhuang and William Zhuk and Barret Zoph},
      year={2024},
      eprint={2303.08774},
      archivePrefix={arXiv},
      primaryClass={cs.CL},
      url={https://arxiv.org/abs/2303.08774}, 
}

@misc{hu2021loralowrankadaptationlarge,
      title={LoRA: Low-Rank Adaptation of Large Language Models}, 
      author={Edward J. Hu and Yelong Shen and Phillip Wallis and Zeyuan Allen-Zhu and Yuanzhi Li and Shean Wang and Lu Wang and Weizhu Chen},
      year={2021},
      eprint={2106.09685},
      archivePrefix={arXiv},
      primaryClass={cs.CL},
      url={https://arxiv.org/abs/2106.09685}, 
}

@inproceedings{Sheng_2025, series={EuroSys ’25},
   title={HybridFlow: A Flexible and Efficient RLHF Framework},
   url={http://dx.doi.org/10.1145/3689031.3696075},
   DOI={10.1145/3689031.3696075},
   booktitle={Proceedings of the Twentieth European Conference on Computer Systems},
   publisher={ACM},
   author={Sheng, Guangming and Zhang, Chi and Ye, Zilingfeng and Wu, Xibin and Zhang, Wang and Zhang, Ru and Peng, Yanghua and Lin, Haibin and Wu, Chuan},
   year={2025},
   month=mar, pages={1279–1297},
   collection={EuroSys ’25} }




\end{document}